\documentclass[runningheads]{llncs}
\usepackage{booktabs}
\usepackage{subfigure}
\usepackage{amssymb}
\usepackage{amsmath}
\usepackage{multirow}
\usepackage{pifont}
\usepackage{xcolor}
\usepackage{subcaption}
\usepackage{bbding}

\usepackage[T1]{fontenc}
\usepackage{graphicx}
\begin{document}
\title{CALAD: Channel-Aware contrastive Learning for multivariate time series Anomaly Detection}
% Channel-Aware Augmentation for contrastive learning in multivariate time series anomaly detection ?
%
\titlerunning{CALAD}
% If the paper title is too long for the running head, you can set
% an abbreviated paper title here
%
% \author{Jaehyeop Hong\orcidID{0009-0004-5418-3014} \and
% Youngbum Hur\orcidID{0000-0002-1113-1730}}
\author{Jaehyeop Hong\orcidID{0009-0004-5418-3014} \and Youngbum Hur\orcidID{0000-0002-1113-1730}\Envelope}
\authorrunning{J. Hong and Y. Hur}
% First names are abbreviated in the running head.
% If there are more than two authors, 'et al.' is used.
%
\institute{Department of Industrial Engineering, Inha University, Incheon, Republic of Korea \\
\email{jaehyeop.hong@inha.edu, youngbum.hur@inha.ac.kr}}
\maketitle              % typeset the header of the contribution

\begin{abstract}
Multivariate time series anomaly detection has become increasingly important in real-world applications, where labeled data are often scarce. Many existing approaches rely on unsupervised learning to model normal patterns, but they often treat all channels equally. This design can dilute anomaly-relevant signals, since not all channels contribute equally to anomaly detection. In this paper, we propose CALAD, a channel-aware contrastive learning framework for multivariate time series anomaly detection. CALAD governs the construction of contrastive samples using estimated channel relevance, allowing the learning process to reflect anomaly semantics rather than generic similarity. Channel relevance is estimated from reconstruction errors of a transformer-based autoencoder and is used to distinguish channels that are more influential to anomalous behaviors. Using this information, we design a channel-wise augmentation strategy in which positive and negative samples are constructed based on whether anomaly-relevant channels are preserved or perturbed. This encourages invariance to changes in irrelevant channels while being sensitive to changes in anomaly-relevant channels. Furthermore, CALAD combines contrastive learning and an auxiliary reconstruction head, allowing the model to learn discriminative representations while retaining normal structures. Experiments on multiple real-world datasets shows that CALAD consistently outperforms existing methods, particularly under distribution shift scenarios. We provide the code for reproducibility at https://github.com/hirundo1218/CALAD
\end{abstract}

\section{Introduction}
With the growing importance of data collection in the era of artificial intelligence, various sensors have been widely utilized, leading to the generation of massive volumes of time series data \cite{darban2024}. 
% A time series, defined as a sequence of data points collected over time, may contain anomalies, which are deviations from expected patterns.
Detecting anomalies in large-scale time series data is critical in real-world scenarios, as it enhances system reliability, supports predictive maintenance, and reduces potential losses. Time series anomaly detection has been applied across various domains, including industrial control systems \cite{mathur2016}, aerospace systems \cite{hundman2018}, and server monitoring \cite{su2019}.

Deep learning–based anomaly detection has attracted growing attention due to its effectiveness in modeling complex and multivariate time series \cite{darban2024}. These methods are often categorized by the availability of labeled data. In practice, time series datasets typically contain very few labeled anomalies or are entirely unlabeled, making unsupervised approaches particularly appealing. Among them, contrastive learning has emerged as a powerful paradigm by constructing supervisory signals directly from data, without requiring anomaly annotations \cite{darban2025,yang2023}.
% In practice, time series datasets typically contain very few labeled anomalies or are entirely unlabeled, making unsupervised approaches particularly appealing \cite{audibert2020,malhotra2015}.

\begin{figure}[t]
\centering
\includegraphics[width=0.7\linewidth]{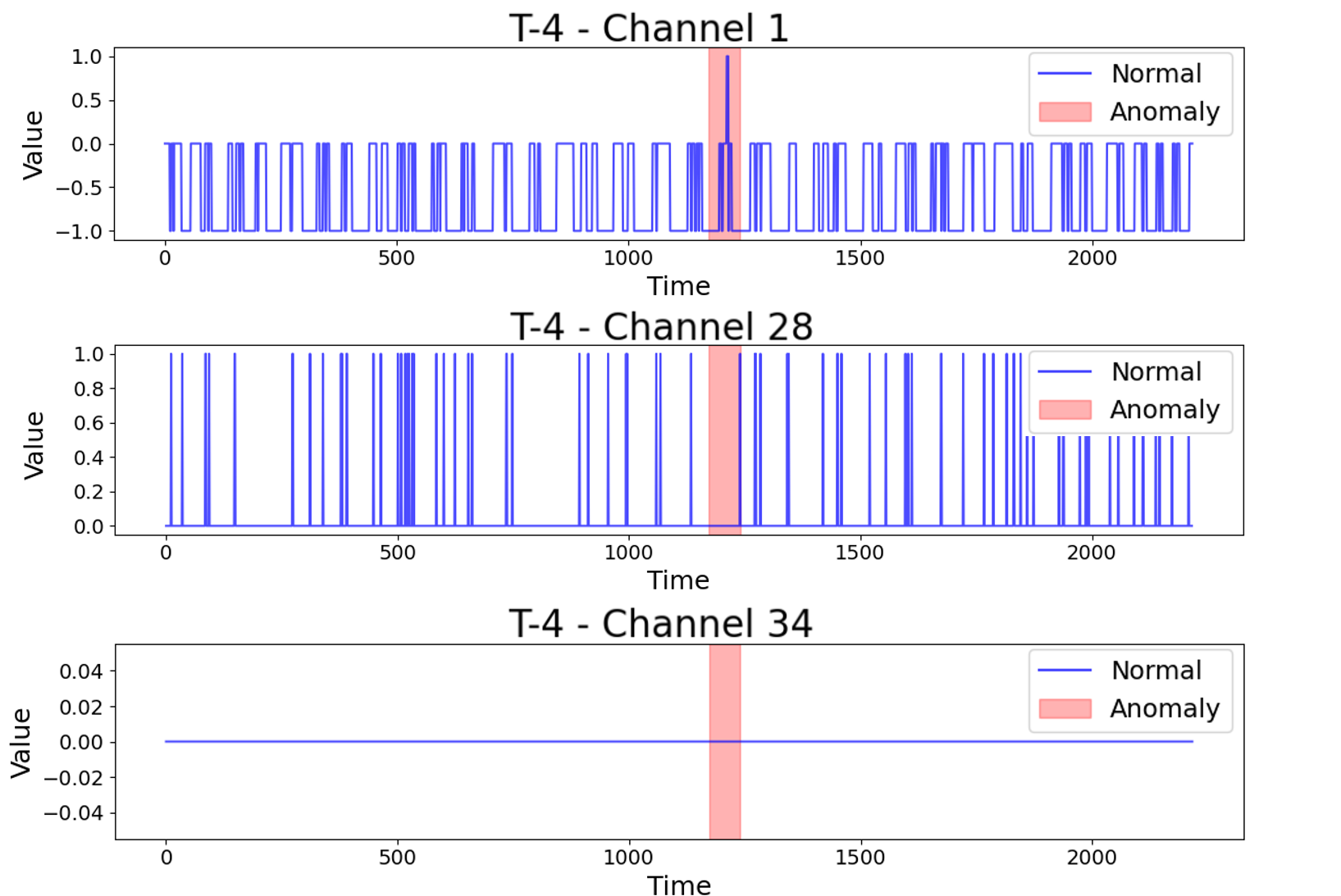}
\caption{An example from the MSL(T-4) dataset showing that each channel exhibits distinct temporal patterns and different degrees of relevance to anomalies.}
\label{fig:1}
\end{figure}

Despite recent progress, unsupervised multivariate time series anomaly detection remains challenging. In practice, not all channels contribute equally to anomaly detection, and treating all channels in the same manner may dilute anomaly-relevant patterns. As illustrated in Fig. \ref{fig:1}, different channels exhibit distinct temporal patterns and varying degrees of association with anomalies, highlighting the necessity of modeling channel-specific roles. 
% Moreover, designing effective data augmentation for contrastive learning in time series is non-trivial, as inappropriate perturbations may distort critical normal structures.

Motivated by these observations, we propose \textbf{CALAD}, a \emph{channel-aware anomaly detection framework} that explicitly incorporates estimated channel relevance into the contrastive learning process. Rather than treating channel importance as a static weighting or attention mechanism, CALAD uses channel relevance to directly govern how contrastive samples are constructed. Specifically, channel relevance is estimated from reconstruction errors, and this information determines whether a channel should be preserved or perturbed when forming positive and negative samples. As a result, contrastive learning is guided by anomaly semantics encouraging invariance to irrelevant channel variations while remaining sensitive to disruptions in anomaly-relevant channels. In addition, CALAD augments contrastive learning with an explicit reconstruction objective, ensuring that normal temporal structures are preserved during representation learning.

Our main contributions are summarized as follows:
\begin{itemize}
\item We propose a channel-aware anomaly detection framework that leverages reconstruction errors as an anomaly indicator to estimate anomaly-relevant channels without using anomaly labels.
\item We introduce a channel-wise contrastive augmentation strategy in which positive and negative samples are constructed based on whether anomaly-relevant channels are preserved or perturbed, aligning contrastive learning with anomaly semantics.
\item We combine contrastive learning with a transformer encoder and an auxiliary reconstruction head, enabling the model to learn discriminative representations while preserving normal temporal structures.
\end{itemize}

\section{Related Works}
\subsection{Time Series Anomaly Detection}
Time series anomaly detection aims to identify observations that deviate from normal system behavior. Early approaches relied on statistical \cite{yamanishi2002}, clustering-based \cite{moshtaghi2014}, and distance-based methods \cite{yeh2016}. More recently, deep learning–based methods have become prevalent, particularly in unsupervised settings where labeled anomalies are scarce. Many recent studies adopt self-supervised learning (SSL) to model normal patterns without anomaly annotations. As summarized in recent surveys \cite{zhang2024}, SSL methods for time series are mainly categorized into generative-based and contrastive-based approaches. Generative methods include autoregressive-based forecasting, such as THOC \cite{shen2020}, and autoencoder-based reconstruction, such as DAEMON \cite{chen2023}, while some methods, such as MTAD-GAT \cite{zhao2020}, combine both paradigms. Contrastive methods, including TS2Vec \cite{yue2022}, DCdetector \cite{yang2023}, and CARLA \cite{darban2025}, have shown strong performance by learning discriminative representations from augmented views. Alongside these learning paradigms, various neural architectures have been explored to capture temporal dependencies in multivariate time series, including RNNs, CNNs, autoencoders, and transformers \cite{darban2024}. In particular, transformer-based models have gained attention due to their ability to capture long-range dependencies, as demonstrated in Anomaly Transformer \cite{xu2022}, TranAD \cite{tuli2022}, and DDDM \cite{wang2023}. However, most methods treat all channels uniformly during representation learning and augmentation, which may limit their ability to capture channel-specific anomaly characteristics.

\subsection{Feature Selection in Time Series}
Feature and channel selection has been widely studied to improve robustness and interpretability in multivariate time series analysis. Many approaches include statistical criteria such as correlation analysis, variance-based measures, and mutual information to identify informative channels \cite{huang2024}, as well as sparsity-inducing regularization techniques, such as LASSO, to automatically select relevant variables by assigning zero coefficients to less important ones \cite{konzen2016}. In anomaly detection, feature selection has been explored to reduce the influence of irrelevant channels \cite{li2024}. However, in unsupervised settings where anomaly labels are unavailable, identifying channels that are more relevant to anomalous behavior remains challenging. Moreover, channel importance is often used only for weighting or selection, without explicitly guiding how representations or augmented views should be constructed during learning. These limitations motivate methods that go beyond channel selection as a preprocessing step, and instead leverage channel relevance to directly guide how representations are learned, especially in contrastive learning for multivariate time series.

\section{Methodology}
\subsection{Problem Statement}
In our task, a multivariate time series $\textbf{X} = \{x_{1}, x_{2}, \cdots, x_{T}\} \in \mathbb{R}^{T \times C}$ represents a sequence of multivariate time points, where each $x_{t} \in \mathbb{R}^{C}$ is an observation at time $t$, $T$ is the number of time steps, and $C$ is the number of channels. The input to the model is a sliding window $w_{i}=\{x_{i}, x_{i+1}, \cdots, x_{i+ws-1}\}$, where $ws$ denotes the window size. For notation, we use $x_{i,c}$ and $w_{i,c}$ to refer to the time point and the window at the $c$-th channel. $N$ is the number of windows. Our task is to detect whether a given multivariate time series window contains an anomaly. The truth label is denoted by $\textit{y}_{i}$, where $\textit{y}_{i}=0$ indicates a normal window, and $\textit{y}_{i}=1$ indicates an abnormal one.

\subsection{Overall Framework}
\begin{figure*}[h]
    \centering
    \includegraphics[width=1.0\linewidth]{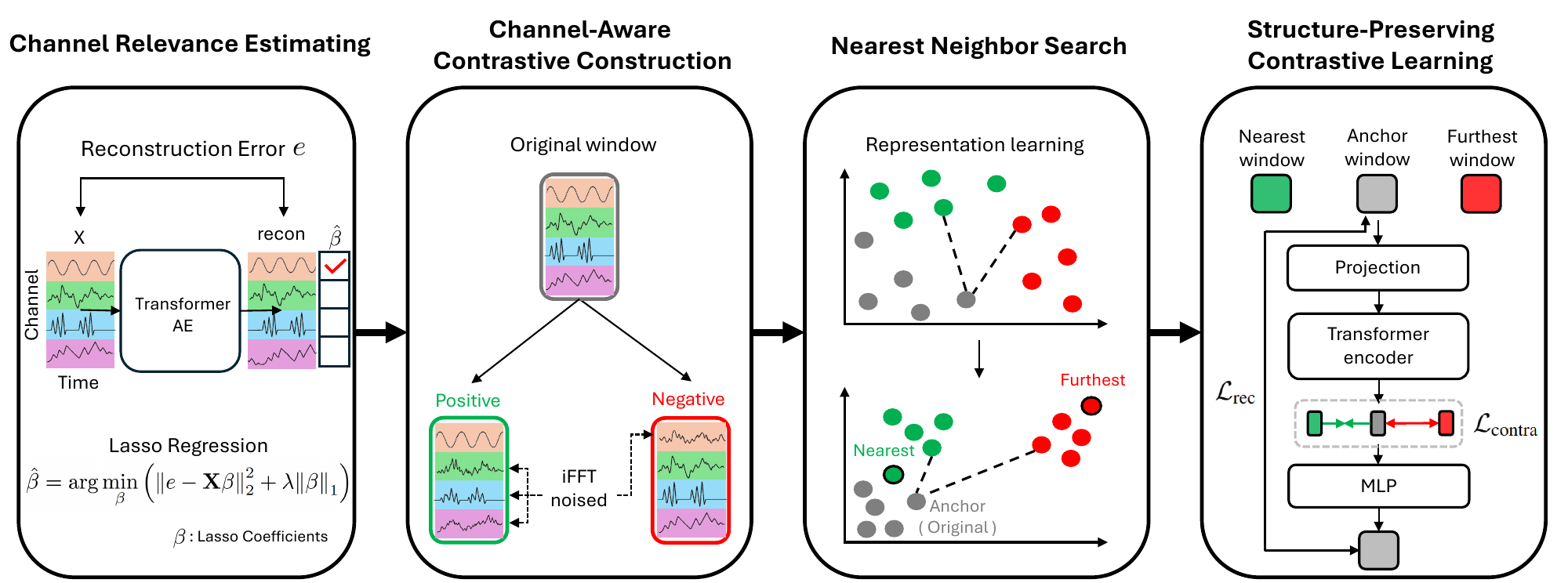}
    \caption{Overall framework of CALAD: We first perform LASSO regression using reconstruction error to estimate channel relevance for contrastive augmentation. Based on this estimated relevance, positive and negative samples are generated via inverse FFT-based augmentation. The samples are embedded into a representation space, where the nearest and farthest neighbors of each anchor (original) window are identified. These three windows are passed through a transformer encoder followed by an MLP-based reconstruction head. The model is jointly optimized using contrastive and reconstruction losses.}
    \label{fig:framework}
\end{figure*}

The overall framework of CALAD is illustrated in Fig. \ref{fig:framework}.  CALAD is designed as a channel-aware contrastive learning pipeline, where estimated channel relevance guides how contrastive samples are constructed and learned. To this end, the framework consists of four stages: Channel Relevance Estimating for Contrastive Augmentation, Channel-Aware Positive and Negative Pair Construction, Nearest Neighbor Search, and Structure-Preserving Contrastive Learning.

\subsection{Channel Relevance Estimating for Contrastive Augmentation}
In this study, we estimate channel relevance to anomalous behaviors by analyzing how strongly each channel contributes to reconstruction error. The goal is not to precisely identify anomaly-relevant channels, but to roughly separate channels that contribute strongly to anomalous behaviors from those that have little influence. This coarse separation is sufficient for guiding channel-aware contrastive augmentation. To achieve this, we use LASSO regression as a estimation tool. LASSO (Least Absolute Shrinkage and Selection Operator) imposes an L1 penalty on regression coefficients $\beta$, which encourages sparsity by shrinking less important coefficients toward zero and thus eliminates irrelevant variables. Given multivariate time series data \textbf{X} and a response variable \textit{y}, the regression coefficients are obtained by solving the following optimization problem:
\begin{equation}
    \hat{\beta} = \textrm{arg}\min_{\beta} \left({\left\| y-\textbf{X}\beta \right \|}_{2}^{2} + \lambda {\left \| \beta \right\|}_{1} \right)
\end{equation}

where $\lambda$ controls the degree of sparsity. Channels with non-zero coefficients in $\hat{\beta}$ are considered more relevant to anomalous behavior. In unsupervised settings, anomaly labels are not available. We therefore use reconstruction error as the response variable \textit{y}. Reconstruction error from reconstruction-based models serves as an anomaly indicator, as it reflects deviations from normal temporal patterns captured during training. Channels whose values explain reconstruction error variations can be interpreted as being more sensitive to anomalous behaviors. Specifically, we train a transformer-based autoencoder to reconstruct data across all channels. Using the trained model, the reconstruction error across channels is computed, where $\hat{x}_{i,c}$ denotes the reconstructed data:
\begin{equation}
     e_{i,c} = \left | x_{i,c} - \hat{x}_{i, c}\right | , \quad y_{i}=\frac{1}{C}\sum_{c=1}^{C}e_{i,c}
\end{equation}

\begin{figure}[t]
    \centering
    \subfigure[LASSO regression coefficients obtained from reconstruction error]{
        \includegraphics[width=0.48\linewidth]{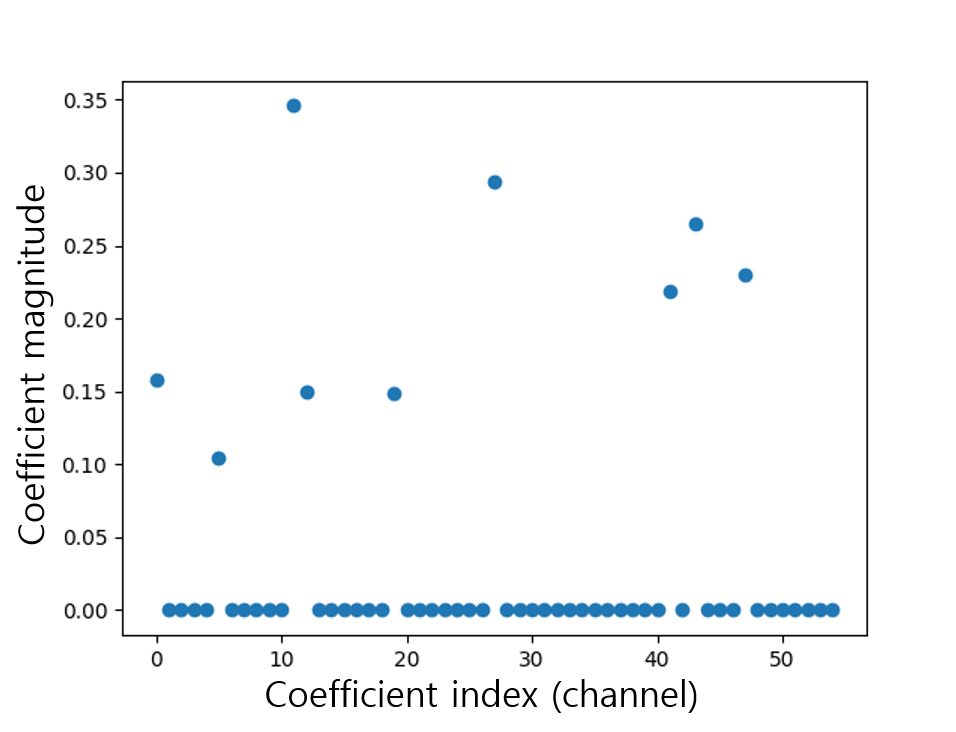}}
    \subfigure[Comparison of channels with zero and non-zero coefficients]{
        \includegraphics[width=0.48\linewidth]{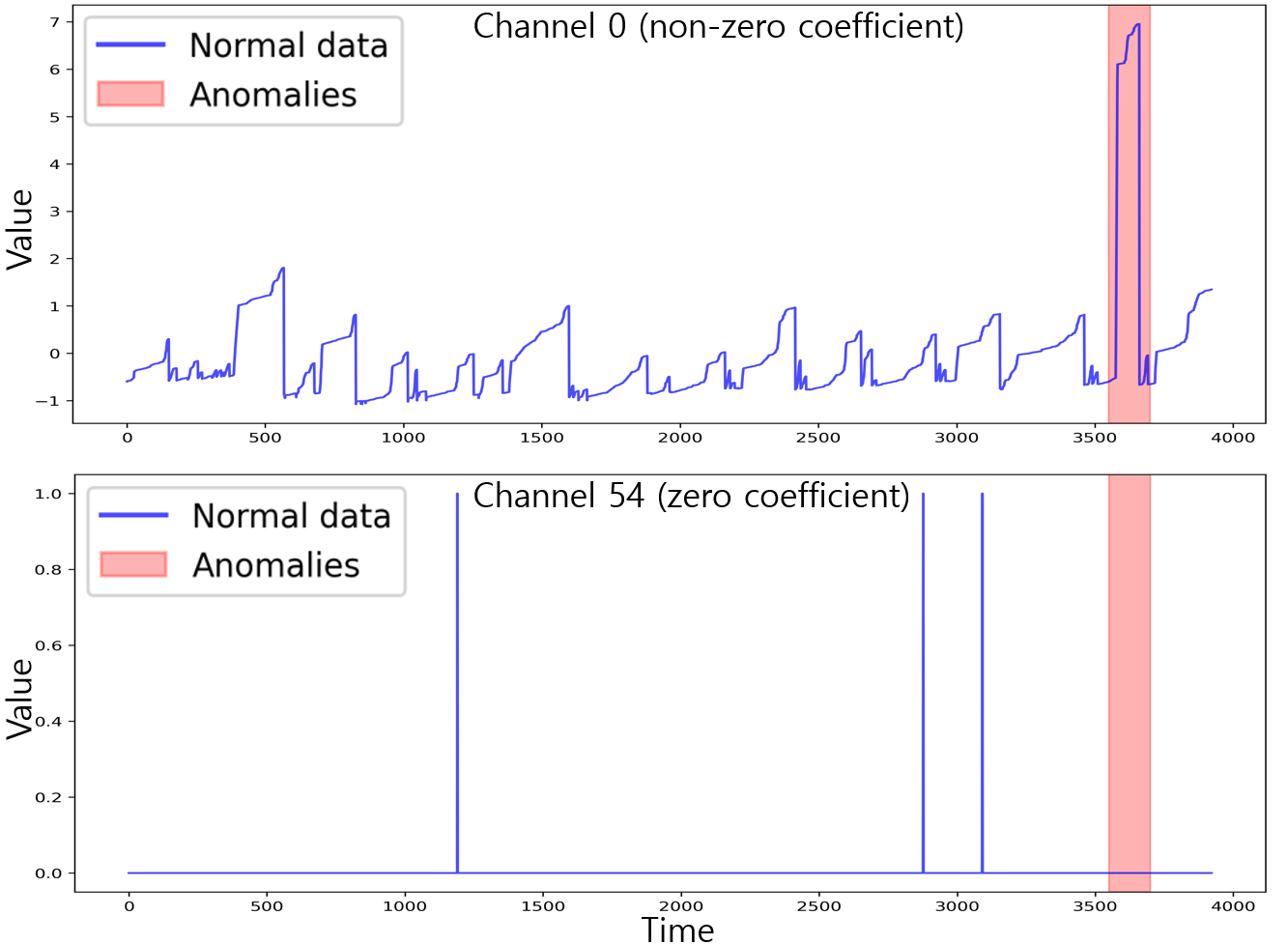}}
    \caption{Channel relevance estimation on the MSL(F-5) dataset.}
    \label{coeff}
\end{figure}

LASSO regression is then performed using the original multivariate input as predictors and the reconstruction error as the response variable. Channels with non-zero coefficients are treated as anomaly-relevant, while those with zero coefficients are treated as anomaly-irrelevant. Fig. \ref{coeff}(a) shows an example of the resulting LASSO coefficients. Only 9 out of 55 channels have non-zero coefficients, while the remaining 46 channels have zero coefficients, indicating that most channels are estimated to be irrelevant to anomalous behaviors. Fig. \ref{coeff}(b) compares channels with non-zero and zero coefficients, illustrating that channels with non-zero coefficients are more associated with anomalous behaviors than those with zero coefficients.

% \begin{figure}[t]
%     \centering
%     \includegraphics[width=0.7\linewidth]{figures/coeff.pdf}
%     \caption{Lasso Regression Coefficients from the MSL(F-5) dataset.}
%     \label{coeff}
% \end{figure}

\subsection{Channel-Aware Positive and Negative Pair Construction}
Time series data require careful handling during augmentation, as inappropriate transformations can distort intrinsic temporal patterns. Applying uniform transformations across all channels, as commonly done in time series augmentation, may unintentionally distort important patterns in anomaly-relevant channels, thereby degrading anomaly-related representations. 

To address this issue, we construct contrastive samples by selectively transforming channels based on their estimated relevance to anomalies. Positive samples are generated by perturbing anomaly-irrelevant channels while preserving anomaly-relevant ones, thereby retaining anomaly-relevant normal patterns and enabling the learning of diverse normal representations. In contrast, negative samples are generated by perturbing anomaly-relevant channels while keeping anomaly-irrelevant channels unchanged, which disrupts anomaly-relevant normal patterns and causes these views to be regarded as anomalous representations. This channel-aware design reduces representation bias from irrelevant channels and encourages the model to focus on anomaly-relevant signals.

We implement the proposed channel-aware construction using inverse FFT perturbations. Channels with non-zero coefficients, estimated as anomaly-relevant, are perturbed only when generating negative samples, while channels with zero coefficients are perturbed only when generating positive samples. As a result, each window yields a triplet set ($w^{anc}_{i}, w^{pos}_{i}, w^{neg}_{i}$) consisting of an anchor (original), a positive, and a negative sample. For each channel $c$ in window $i$, inverse FFT augmentation is selectively applied according to the channel label. The augmentation process is defined as follows:
\begin{equation}
\begin{aligned}
\text{label}(c) = 
\begin{cases}
    0 & \text{if zero coefficient}   \\
    1 & \text{otherwise} \\
    \end{cases} \\
\end{aligned}
\end{equation}
\begin{equation}
\begin{aligned}
w^{pos}_{i} &= \text{Augment}_{\text{iFFT}}(w^{anc}_{i,c}, \{c \mid \text{label}(c) = 0\}) \\
w^{neg}_{i} &= \text{Augment}_{\text{iFFT}}(w^{anc}_{i,c}, \{c \mid \text{label}(c) = 1\}) \\
\end{aligned}
\end{equation}

\subsection{Nearest Neighbor Search}
To organize channel-aware triplets into a globally consistent representation space, we encode each triplet using an encoder and optimize it with a triplet loss. Since positive and negative samples are constructed based on channel relevance to anomalies, this objective aligns representations according to anomaly semantics rather than generic similarity. Specifically, the anchor and positive windows are pulled closer, while the anchor and negative windows are pushed farther apart in the representation space. Given a triplet sets $\mathcal{T} = \{(w_i^{anc}, w_i^{pos}, w_i^{neg})\}_{i=1}^N$, each window is encoded using the encoder $f(\cdot)$ to obtain latent representations, and the encoder is trained using the following triplet loss:
\begin{equation}
    z_i^{anc} = f(w_i^{anc}), \quad z_i^{pos} = f(w_i^{pos}), \quad z_i^{neg} = f(w_i^{neg})
\end{equation}
\begin{equation}
\mathcal{L}_{\textit{T}} = \frac{1}{N} \sum_{i=1}^N 
\max \Big(
    \| z_i^{anc} - z_i^{pos} \|_2^2 
    - \| z_i^{anc} - z_i^{neg} \|_2^2 
    + \alpha,\; 0
\Big)
\end{equation}

where $\alpha$ is the margin hyperparameter.
%that encourages separation between positive and negative pairs in the representation space.
After training, the resulting representation space allows us to identify meaningful neighbors for each anchor window. Let $\mathcal{Z}$ denote the set of all window representations. For each anchor, we retrieve its nearest and furthest neighbors based on Euclidean distance:
\begin{align}
    \text{Nearest}(z_i^{anc}) &= \arg\min_{z \in \mathcal{Z}} \| z_i^{anc} - z \|_2^2 \\
    \text{Furthest}(z_i^{anc}) &= \arg\max_{z \in \mathcal{Z}} \| z_i^{anc} - z \|_2^2
    % \setminus \{z_i^a\}
\end{align}

\subsection{Structure-Preserving Contrastive Learning}
The nearest and furthest neighbors identified in the previous stage define a meaningful neighborhood for each anchor window. While the channel-aware construction provides meaningful positive and negative views, effective anomaly detection further requires learning representations that preserve the structural features of normal data. To this end, we construct new triplet sets $\{(w_i^{anc}, w_i^{near}, w_i^{fur})\}_{i=1}^N$ and jointly perform contrastive learning and reconstruction. Each window is projected through a linear layer and processed by a transformer encoder. The resulting embeddings $(z_i^{anc}, z_i^{near}, z_i^{fur})$ are used for contrastive learning, which encourages the anchor to be close to its nearest neighbor and far from its furthest neighbor, reinforcing semantic consistency in the representation space.

However, contrastive learning alone mainly focuses on relative separation and does not explicitly preserve the original normal structures. To address this, we introduce an auxiliary reconstruction objective applied only to the anchor window. Specifically, the anchor embedding is passed through an MLP-based reconstruction head $f_{\text{mlp}}(\cdot)$ to reconstruct the original window. This design avoids reconstructing perturbed views while explicitly constraining the representation to retain fine-grained normal patterns. By jointly optimizing contrastive and reconstruction objectives, the model learns representations that are discriminative to anomalies while preserving normal temporal structures. The contrastive loss is computed based on the similarity between embeddings, defined as $\text{sim}(z_{i},z_{j})=p_{i} \cdot p_{j}$, where $p_{i}=\text{Softmax}(z_{i})$ and BCE denotes binary cross entropy. The reconstruction loss is defined as the mean squared error between the anchor window and its reconstruction $\hat{w}^{anc}_{i}=f_{\text{mlp}}(z^{anc}_{i})$. The overall objective is defined as follows:
\begin{equation}
\begin{aligned}
\mathcal{L}_{\text{contra}} = \frac{1}{N} \sum_{i=1}^N \Big(\text{BCE}(\text{sim}(z^{anc}_{i},z^{near}_{i}),1)-\text{BCE}(\text{sim}(z^{anc}_{i},z^{fur}_{i}),0)\Big)
\end{aligned}
\end{equation}
\begin{equation}
\mathcal{L}_{\text{rec}} = \frac{1}{N} \sum_{i=1}^N \|w^{anc}_{i} - \hat{w}^{anc}_{i}\|_2^2
\end{equation}
\begin{equation}
\mathcal{L}_{\text{Total}} = \mathcal{L}_{\text{contra}} + \mathcal{L}_{\text{rec}}
% \lambda \cdot
\end{equation}

\subsection{Anomaly Scoring}
% CALAD assigns each window to one of $K=\{1,\dots,k\}$ classes using a softmax function, where $k=10$ is empirically chosen to capture diverse normal patterns beyond a binary classification. The softmax output produces a probability distribution $p(x)$, where $p^{k}(x)$ denotes the probability that input $x$ belongs to class $k$. During training, each anchor window is assigned to the class with the highest probability. The normal class is defined as the class to which the largest number of training windows are assigned, and is assumed to represent the most common normal pattern. For a test window $w_i$, CALAD computes $p^{\text{normal}}(w_i)$, the probability of belonging to the normal class. If this probability is the highest among all classes, the window is labeled as normal, otherwise, it is labeled as abnormal. The anomaly score is defined as the inverse of this probability, with higher scores indicating more anomalous behavior. The anomaly label and score are defined as follows:

CALAD assigns each window to one of $K=\{1,\dots,k\}$ latent classes using a softmax function, where $k=10$ is empirically chosen to balance model expressiveness and robustness. The use of multiple latent classes improves the representational capacity of the model in an unsupervised setting. The softmax output produces a probability distribution $p(x)$, where $p^{k}(x)$ denotes the probability that input $x$ belongs to class $k$. We define the normal class as the one with the highest assignment frequency in the training data, based on the assumption that normal patterns are dominant. For a test window $w_i$, CALAD computes $p^{\text{normal}}(w_i)$, the probability of belonging to the normal class. If this probability is the highest among all classes, the window is labeled as normal, otherwise, it is labeled as abnormal. The anomaly score is defined as the inverse of this probability, with higher scores indicating more anomalous behavior. The anomaly label and score are defined as follows:

\begin{equation}
\text{Anomaly Label }(w_{i}) =
\begin{cases}
0, & \text{if } p^{\text{normal}}(w_i) \text{ is the highest},\\
1, & \text{otherwise}
\end{cases}
\end{equation}
\begin{equation}
        \text{Anomaly Score }(w_i)=1-p^{\text{normal}}(w_i)
\end{equation}

\section{Experiments}
\subsection{Datasets}
We evaluate CALAD on four widely used multivariate time series anomaly detection datasets, summarized in Table \ref{tab:datasets}. MSL and SMAP \cite{hundman2018} are spacecraft monitoring datasets collected from a NASA satellite and the Curiosity rover, respectively. SMD \cite{su2019} is a large-scale server monitoring dataset, and SWaT \cite{mathur2016} is an industrial control system dataset from a water treatment plant.
\begin{table}
    \centering
    \caption{Statistics of the datasets. Entity denotes the number of distinct time series. Dim is short for dimension and AR is short for the anomaly rate.}
    \begin{tabular}{ccccc|ccccccccccccccccccc}
        \toprule
        && Dataset && &&&& \#Train &&&& \#Test\phantom{z} &&& Entity &&& Dim &&& AR(\%) && \\
        \midrule
        && MSL  && &&&& 58,317  &&&& 73,729  &&& 27 &&& 55 &&& 10.48 && \\
        && SMAP && &&&& 135,183 &&&& 427,617 &&& 55 &&& 25 &&& 12.83 && \\
        && SMD  && &&&& 708,405 &&&& 708,420 &&& 28 &&& 38 &&& 4.16 && \\
        && SWaT && &&&& 495,000 &&&& 449,919 &&& 1  &&& 51 &&& 12.14 && \\
        \bottomrule
    \end{tabular}
    \label{tab:datasets}
\end{table}

\subsection{Experimental Setup}
\paragraph{\textbf{Baselines.}}
We compare CALAD with representative anomaly detection methods, including LSTM-VAE \cite{park2018}, OmniAnomaly \cite{su2019}, MSCRED \cite{zhang2019}, MTAD-GAT \cite{zhao2020}, THOC \cite{shen2020}, Anomaly Transformer (AnomTran) \cite{xu2022}, UAE \cite{garg2021}, TranAD \cite{tuli2022}, TS2Vec \cite{yue2022}, DCdetector \cite{yang2023}, TimesNet \cite{wu2022}, EdgeConvFormer \cite{liu2024}, and CARLA \cite{darban2025}. In addition, we compare it with random anomaly scores (Random).

\paragraph{\textbf{Evaluation Metrics.}}
We evaluate performance using precision (P), recall (R), F1-score (F1), and area under the precision recall curve (AU-PR). Note that we do not use Point Adjustment (PA) in our evaluation. Recent studies \cite{ghorbani2024,wang2023} have demonstrated that PA can lead to faulty performance evaluations, where PA uses true labels from the test datasets to adjust the outputs of models. It is known that using PA can result in state-of-the-art performance even with random guess or randomly initialized non-trained models \cite{darban2025}, making it impossible to conduct a fair comparison and assess the effectiveness of the models. Therefore, following previous work \cite{darban2025}, we use original F1 as an evaluation metric instead of PA.

\paragraph{\textbf{Implementation Details.}} We use a window size of 200 with a stride of 1 for all datasets. During the training, the Adam optimizer is used with a batch size of 50. For channel relevance estimation, the regularization parameter of LASSO is set to 0.001, which was empirically chosen to achieve a balance between sparsity and stability in channel selection. In the nearest neighbor search stage, the ResNet is trained for 30 epochs with learning rate of 0.001. In the structure-preserving contrastive learning stage, our model is trained for 50 epochs with learning rate of 0.01. All experiments are conducted on an NVIDIA RTX A6000 GPU.

\subsection{Main Results}
Table \ref{tab:main} reports the anomaly detection performance of CALAD and baseline methods on four benchmark datasets. Overall, CALAD achieves the best performance on three datasets and the second-best performance on the remaining one, with only a marginal gap. On average, CALAD yields an 11\% relative improvement in F1-score over the strongest baseline methods. In particular, CALAD achieves notable F1-score improvements of 31\% and 13\% on the MSL and SMAP datasets, respectively. On the SMD dataset, which contains a large volume of data with a relatively low anomaly rate, CALAD achieves the best F1-score, demonstrating its robustness in large-scale multivariate settings. On the SWaT dataset, CALAD attains the second-best performance, with an F1-score within 1\% of the best method, as the benefit of learning channel-semantic diversity is less pronounced in single entity settings.
\begin{table*}
    \centering
    \caption{Anomaly detection results on four benchmark datasets. AU-PR and F1 are reported in percentage (\%). The best F1-score is highlighted in bold, and the second-best F1-score is underlined.}
    \resizebox{1\linewidth}{!}{
    \begin{tabular}{ll|ccc|ccc|ccc|ccc}
        \toprule
        \toprule
        \multicolumn{2}{c}{Dataset} & \multicolumn{3}{c}{MSL} & \multicolumn{3}{c}{SMAP} & \multicolumn{3}{c}{SMD} & \multicolumn{3}{c}{SWaT} \\  
        \cmidrule{3-14}
        \multicolumn{2}{c}{Metrics} & \phantom{z}AU-PR & \phantom{z}F1 && \phantom{z}AU-PR & \phantom{z}F1 && \phantom{z}AU-PR & \phantom{z}F1 && \phantom{z}AU-PR & \phantom{z}F1 &\\
        \midrule
        \midrule
        Random         && 17.20 & 29.36 && 14.00 & 30.28 &&  8.90 & 17.31 && 12.90 & 21.66 &\\
        LSTM-VAE       && 28.50 & 40.74 && 25.80 & 43.70 && 39.50 & 29.80 && 68.50 & 72.37 &\\
        OmniAnomaly    && 22.01 & 18.45 && 22.69 & 18.81 && 41.66 & 35.19 && 21.35 & 24.25 &\\
        MSCRED         && 25.56 & 21.59 && 24.49 & 14.87 && 41.74 & 35.37 && 22.98 & 13.19 &\\
        MTAD-GAT       && 33.50 & 47.34 && 33.90 & 51.86 && 40.10 & 34.73 &&  9.50 & 24.23 &\\
        THOC           && 23.90 & 30.95 && 19.50 & 32.73 && 10.70 & 16.79 && 53.70 & 63.80 &\\
        AnomTrans      && 23.60 & 34.49 && 26.40 & 40.74 && 27.30 & 30.43 && 68.10 & \textbf{73.76} &\\
        UAE            && 17.41 &  6.38 && 15.90 & 14.75 && 39.55 & 23.79 && 43.44 &  8.34 &\\
        TranAD         && 27.80 & 42.83 && 28.70 & 47.16 && 41.20 & 36.09 && 19.20 & 31.03 &\\
        TS2Vec         && 13.20 & 29.93 && 14.80 & 37.12 && 11.30 & 17.28 && 13.60 & 26.11 &\\
        DCdetector     && 12.90 & 22.70 && 12.40 & 27.53 &&  4.30 &  8.28 && 12.60 & 21.66 &\\
        TimesNet       && 28.30 & 35.78 && 20.80 & 40.19 && 38.50 & 33.85 &&  8.30 & 21.66 &\\
        EdgeConvFormer && 32.47 & 34.04 && 31.24 & 17.66 && 39.67 & 27.28 && 70.68 & 72.75 &\\
        CARLA          && 50.10 & \underline{52.27} && 44.80 & \underline{52.92} && 50.70 & \underline{51.14} && 68.10 & 72.09 &\\
        \midrule
        CALAD          && 61.20 & \textbf{68.24} && 52.30 & \textbf{60.00} && 51.10 & \textbf{51.53} && 70.55 & \underline{72.87} &\\
        \bottomrule
        \bottomrule
    \end{tabular}
    }
    \label{tab:main}
\end{table*}

\begin{figure}[t]
    \centering
    \includegraphics[width=0.8\linewidth]{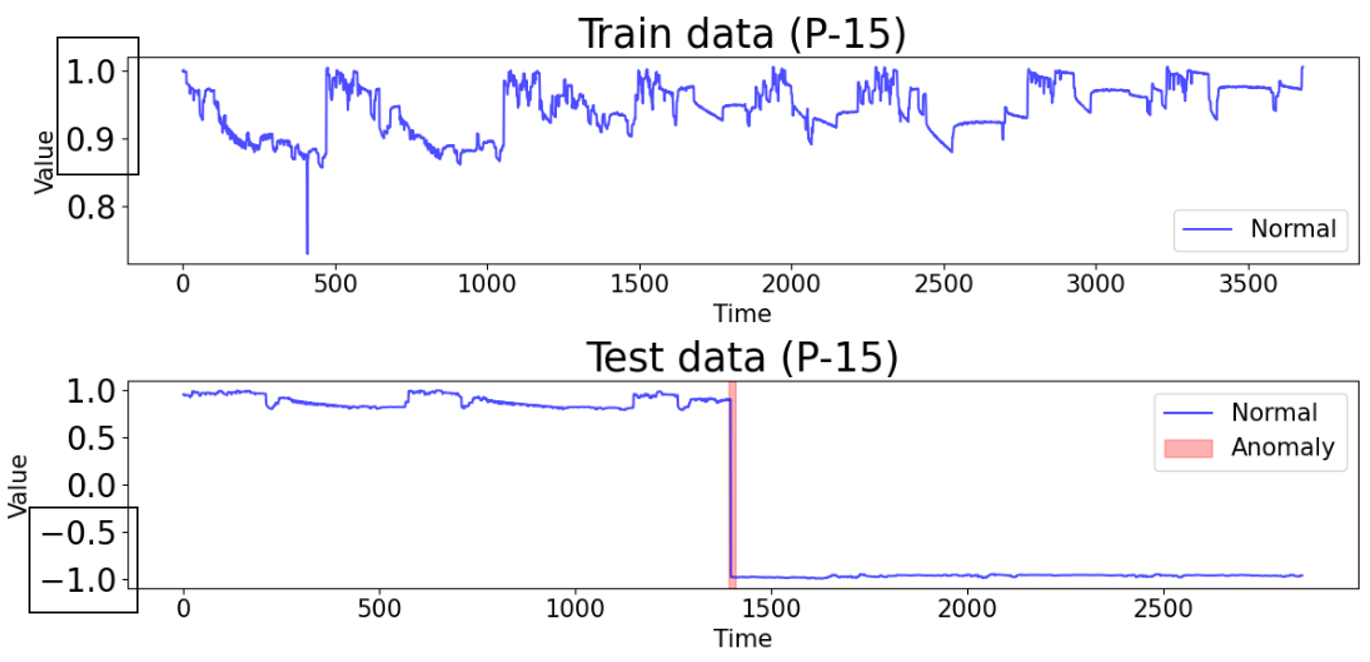}
    \caption{The distribution shift in the MSL(P-15) dataset. Normal patterns with values near -1.0 appear in the test data but are not observed during training.}
    \label{fig:3}
\end{figure}

The strong performance on MSL and SMAP is closely related to their severe distribution shift, as illustrated in Fig. \ref{fig:3}, where normal patterns unseen during training appear in the test data. Such shifts make it difficult for many baselines to correctly recognize unseen normal patterns. CALAD alleviates this issue through positive samples generated by channel-aware augmentation, which preserves anomaly-relevant channels while varying irrelevant ones, enabling the model to learn diverse normal representations and improve robustness under distribution shift. As a result, CALAD better recognizes unseen normal patterns, leading to improved detection performance.

In addition, CALAD consistently outperforms baseline contrastive methods such as TS2Vec \cite{yue2022}, DCdetector \cite{yang2023}, and CARLA \cite{darban2025}. While these methods primarily emphasize representation separation through contrastive learning, CALAD additionally incorporates a reconstruction objective to explicitly preserve normal temporal structures. This complementary constraint contributes to more stable representations of normal data, supporting improved anomaly detection performance. 
% The benefit of this design is further supported by the ablation results, where removing the reconstruction objective consistently degrades performance, indicating its complementary role to contrastive learning.

\subsection{Ablation Studies}
We conduct ablation studies on the MSL dataset, a challenging benchmark characterized by high dimensionality and strong distribution shift, to better examine the effectiveness of each component under realistic and difficult conditions.

\paragraph{\textbf{Effect of Augmentation Strategies.}}
We first analyze the effect of channel-aware construction. As shown in Table \ref{tab:distribution}, results on MSL(P-15) demonstrate that channel-wise augmentation outperforms all-channel augmentation under severe distribution shift. By preserving anomaly-relevant channels while perturbing only anomaly-irrelevant ones when constructing positive samples, By preserving anomaly-relevant channel while perturbing only anomaly-irrelevant ones when constructing positive samples, the model learns more diverse normal representations and becomes more robust to unseen normal patterns. In contrast, uniformly applying Gaussian noise to all channels tends to distort important patterns, resulting in degraded performance.
\begin{table}[t]
    \centering
    \caption{Effects of augmentation strategies, noise types, and backbone architectures on the MSL dataset.}
    \begin{tabular}{cccccccccccccccccc}
        \toprule
        && Entity &&&& Method                                             &&& P     &&& R     &&& F1             &&\\
        \midrule
        &&        &&&& \multicolumn{1}{c}{\textbf{\textit{Augmentation}}} &&&       &&&       &&&                &&\\
        && P-15   &&&& All-channel                                        &&& 99.10 &&& 50.00 &&& 66.47          &&\\
        % Gaussian noise    &&       &&       &&       \\
        &&        &&&& Channel-wise                                       &&& 99.99 &&& 64.55 &&& \textbf{78.45} &&\\
        \midrule
        &&        &&&& \multicolumn{1}{c}{\textbf{\textit{Noise}}}        &&&       &&&       &&&                &&\\
        &&        &&&& Anomaly Injection                                  &&& 51.89 &&& 74.49 &&& 61.12          &&\\
        && \multirow{3}*{All-Entities} &&&& iFFT                           &&& 64.69 &&& 72.19 &&& \textbf{68.24} &&\\
        \cmidrule{7-18}
        &&        &&&& \multicolumn{1}{c}{\textbf{\textit{Backbone}}}     &&&       &&&       &&&                &&\\
        &&        &&&& ResNet                                             &&& 43.80 &&& 74.13 &&& 55.07          &&\\
        &&        &&&& Transformer                                        &&& 52.49 &&& 73.98 &&& 61.41          &&\\
        &&        &&&& Transformer-MLP                                    &&& 64.69 &&& 72.19 &&& \textbf{68.24} &&\\
        \bottomrule
    \end{tabular}
    \label{tab:distribution}
\end{table}

\begin{table}[t]
    \centering
    \caption{Ablation on the effect of loss function components on the MSL dataset.}
    \begin{tabular}{ccccccccccccccccccccccc}
        \toprule
        &&& $\mathcal{L}_{\text{contra}}$ &&&& $\mathcal{L}_{\text{recon}}$ &&&&& P     &&&& R     &&&& F1             &&\\
        \midrule
        &&& --                             &&&& \ding{51}                    &&&&& 26.83 &&&& 79.74 &&&& 40.15          &&\\
        &&& \ding{51}                      &&&& --                           &&&&& 49.61 &&&& 70.60 &&&& 58.27          &&\\
        &&& \ding{51}                      &&&& \ding{51}                    &&&&& 64.69 &&&& 72.19 &&&& \textbf{68.24} &&\\
        \bottomrule
    \end{tabular}
    \label{tab:lossfuction}
\end{table}

We further compare different augmentation noise strategies in Table \ref{tab:distribution}. The anomaly injection method \cite{darban2025}, which generates negative samples by randomly injecting synthetic abnormal patterns, shows relatively lower performance. While this approach aims to simulate realistic anomalies, such injections may distort important patterns. In contrast, the proposed inverse FFT–based perturbation selectively modifies channel-specific frequency components, avoiding unnecessary distortion of important patterns. These results highlight the importance of meaningful perturbations in contrastive learning for anomaly detection.

\paragraph{\textbf{Effect of Architecture.}}
To evaluate the effectiveness of CALAD’s architectural design, we compare different backbones, including ResNet, a Transformer encoder, and a Transformer encoder combined with an MLP-based reconstruction head. As shown in Table \ref{tab:distribution}, the architecture incorporating both the transformer encoder and the reconstruction head achieves the highest F1-score. While the Transformer encoder enhances discriminative representations through contrastive learning, the additional reconstruction head encourages the model to better capture and preserve normal structural patterns. This improved modeling of normal data reduces false positive detections, which is reflected in the higher precision observed in the results.

\begin{figure}[t]
    \centering
    \subfigure[Without reconstruction]{
        \includegraphics[width=0.48\linewidth]{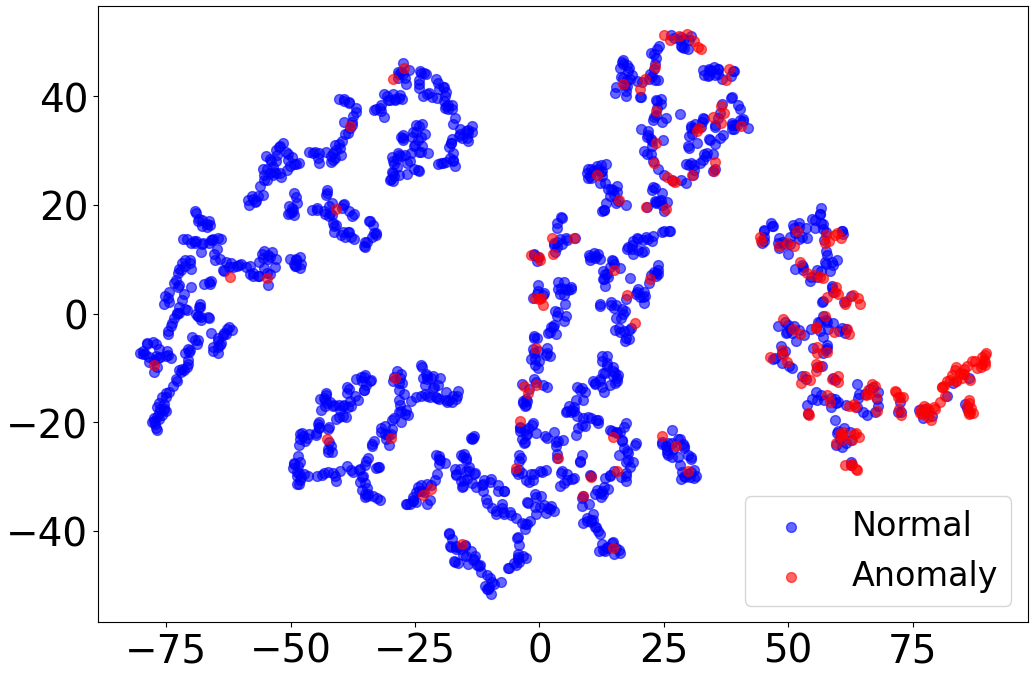}}
    \subfigure[With reconstruction]{
        \includegraphics[width=0.48\linewidth]{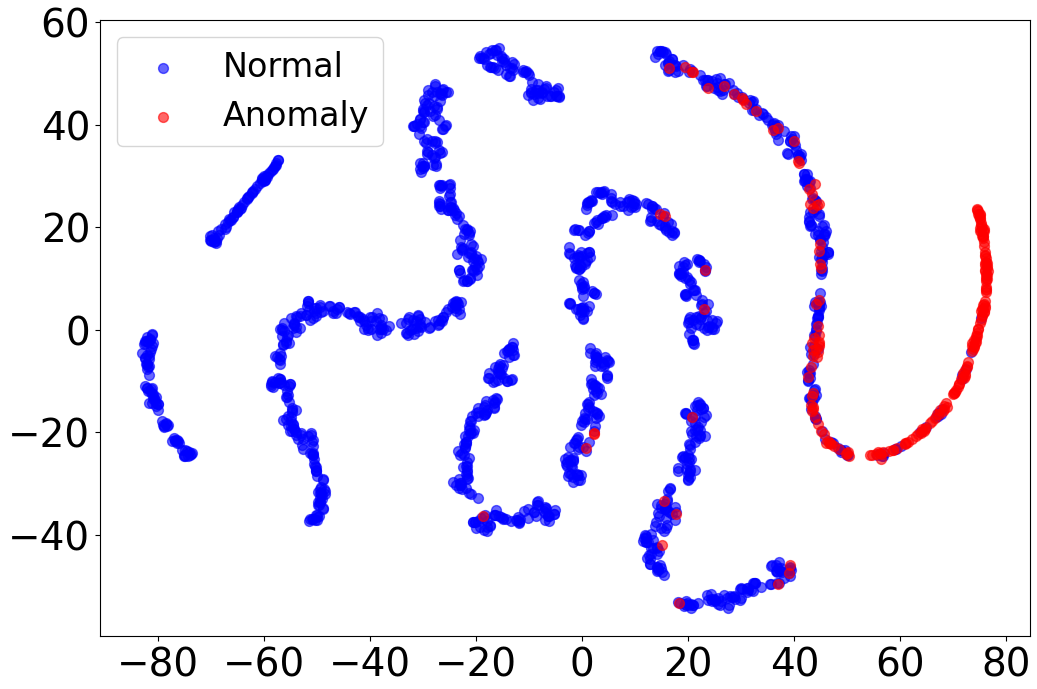}}
    \subfigure[Without reconstruction]{
        \includegraphics[width=0.48\linewidth]{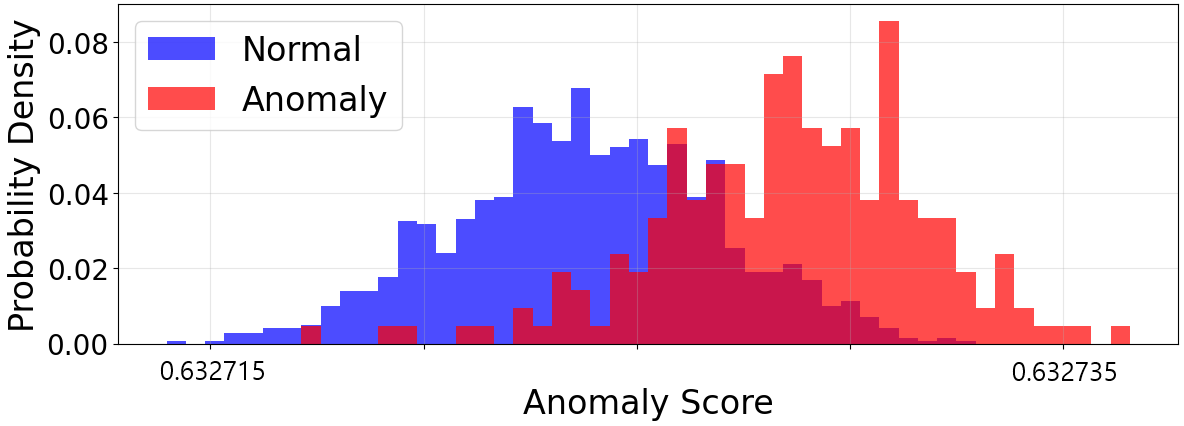}}
    \subfigure[With reconstruction]{
        \includegraphics[width=0.48\linewidth]{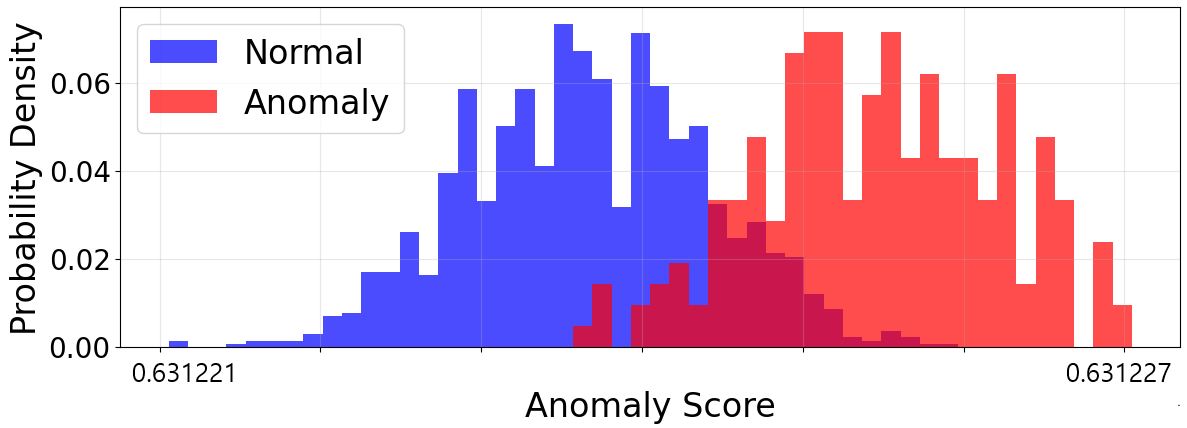}}
        
    \caption{Visualization of t-SNE embeddings and anomaly score distributions on the MSL(S-2) dataset, comparing models with and without reconstruction.}
    \label{fig:tsne}
\end{figure}

\paragraph{\textbf{Effect of Loss Components.}}
We analyze the contribution of each loss component by comparing contrastive loss only, reconstruction loss only, and their combination. As reported in Table \ref{tab:lossfuction}, jointly optimizing both losses achieves the best performance on the MSL dataset. As illustrated in Fig. \ref{fig:tsne}, incorporating reconstruction enables the model to better preserve normal structural patterns, resulting in a clearer separation between normal and anomalous samples in the anomaly score distribution. This effect is quantitatively confirmed in Table \ref{tab:recon}, where adding the reconstruction objective leads to a relative improvement of 33\% in F1-score. These results demonstrate that contrastive and reconstruction losses provide complementary benefits when combined.
\begin{table}[t]
    \centering
    \caption{Effect of the reconstruction objective on the MSL(S-2) dataset.}
    \begin{tabular}{ccccccccccccccccc}
        \toprule
        && Methods                &&&& P     &&&& R     &&&& F1             &&\\
        \midrule
        && Without reconstruction &&&& 51.81 &&&& 61.43 &&&& 56.21          &&\\
        && With reconstruction    &&&& 78.53 &&&& 71.43 &&&& \textbf{74.81} &&\\
        \bottomrule
    \end{tabular}
    \label{tab:recon}
\end{table}

\section{Conclusion}
In this paper, we presented CALAD, a multivariate time series anomaly detection framework that incorporates channel-aware modeling into contrastive representation learning. Rather than treating channels uniformly, CALAD embeds anomaly-related channel information into the contrastive learning process, allowing the construction of positive and negative samples to be guided by estimated channel relevance. CALAD employs reconstruction error as an anomaly indicator to estimate channel relevance, which is used to design channel-aware contrastive augmentation. In addition, CALAD jointly optimizes contrastive learning with a transformer encoder and an MLP reconstruction head, allowing representations to remain discriminative while preserving normal temporal structures. Experimental results on multiple benchmark datasets demonstrate that CALAD consistently improves anomaly detection performance, particularly under distribution shift scenarios. While the framework consists of multiple components, it suggests future directions toward more efficient designs without sacrificing performance.

\begin{credits}
\subsubsection{\ackname}
This work was supported by the Institute of Information and Communications Technology Planning and Evaluation (IITP) Grant funded by Korean Government through Ministry of Science and ICT (MSIT) (XVoice: Multi-Modal Voice Meta Learning) under Grant
2022-0-00641.

% \subsubsection{\discintname}
% The authors have no competing interests to declare that are relevant to the content of this article.
\end{credits}

% \subsubsection{Acknowledgements}
% % This work was supported in part by the Institute of Information and Communications Technology Planning and Evaluation (IITP) Grant funded by Korean Government through Ministry of Science and ICT (MSIT) (XVoice: Multi-Modal Voice Meta Learning) under Grant
% % 2022-0-00641, and in part by the Inha University Research Grant.
% This work was supported by the Institute of Information and Communications Technology Planning and Evaluation (IITP) Grant funded by Korean Government through Ministry of Science and ICT (MSIT) (XVoice: Multi-Modal Voice Meta Learning) under Grant
% 2022-0-00641.

% ---- Bibliography ----
%
% BibTeX users should specify bibliography style 'splncs04'.
% References will then be sorted and formatted in the correct style.
% %
\bibliographystyle{splncs04}
\bibliography{ref}

\end{document}